\documentclass[letterpaper, 10 pt, journal, twoside]{IEEEtran}
\ifCLASSINFOpdf
\else
\fi
\usepackage[ruled, linesnumbered]{algorithm2e}
\usepackage{comment}
\usepackage{tikz}
\usetikzlibrary{positioning,fit,backgrounds,calc}
\usepackage{subfloat}
\usepackage{subfig}
\usepackage{diagbox}
\usepackage{epstopdf}
\usepackage{pifont}
\usepackage{amsmath}
\usepackage{multirow}
\usepackage{url}
\usepackage{verbatim}
\usepackage{footmisc}
\usepackage{color}
\usepackage{tabularx}
\usepackage{float}
\usepackage{booktabs}
\usepackage{makecell}
\usepackage{bm}
\usepackage[utf8]{inputenc}
\usepackage{graphicx}
\usepackage{amssymb}
\usepackage{threeparttable}
\usepackage{hyperref}
\usepackage[noadjust]{cite}

\begin{document}
\bibliographystyle{IEEEtran}
%

\title{osmAG-LLM: Zero-Shot Open-Vocabulary Object Navigation via Semantic Maps and Large Language Models Reasoning}
%
%
%

\author{Fujing Xie$^{1}$, S\"{o}ren Schwertfeger$^{1}$, and Hermann Blum$^{2}$%

\thanks{Manuscript received: July, 15, 2025; Revised October, 18, 2025; Accepted December, 2, 2025.}
\thanks{This paper was recommended for publication by Editor Markus Vincze upon evaluation of the Associate Editor and Reviewers' comments.
This work was supported by the Shanghai Frontiers Science Center of Human-centered Artificial Intelligence and German Federal Ministry of Education and Research (BMBF) in the project “Robotics Institute Germany”, grant No. 16ME0999.
The work was made possible with the support of a scholarship from the German Academic Exchange Service (DAAD).
The experiments of this work were supported by the core facility Platform of Computer Science and Communication, SIST, ShanghaiTech University.} 

\thanks{$^{1}$Fujing Xie and S\"{o}ren Schwertfeger are with the Key Laboratory of Intelligent Perception and Human-Machine Collaboration -- ShanghaiTech University, Ministry of Education, China.
        {\tt\footnotesize {xiefj, soerensch}@shanghaitech.edu.cn}}%
\thanks{$^{2} $Hermann Blum is with Robot Perception and Learning Lab, University of Bonn and Lamarr Institute for ML and AI, 53115 Bonn, Germany.
        {\tt\footnotesize blumh@uni-bonn.de}}%
\thanks{Digital Object Identifier (DOI): see top of this page.}
}
%
%

\markboth{IEEE Robotics and Automation Letters. Preprint Version. Accepted December, 2025}
{Xie \MakeLowercase{\textit{et al.}}: osmAG-LLM: Zero-Shot Open-Vocabulary Object Navigation via Semantic Maps and Large Language Models Reasoning} 

%



\maketitle 

\begin{abstract}
Recent open-vocabulary robot mapping methods
enrich dense geometric maps with pre-trained visual-language features, achieving a high level of detail and guiding robots to find objects specified by open-vocabulary language queries. While the issue of scalability for such approaches has received some attention, another fundamental problem is that high-detail object mapping quickly becomes outdated, as objects get moved around a lot. In this work, we develop a mapping and navigation system for object-goal navigation that, from the ground up, considers the possibilities that a queried object can have moved, or may not be mapped at all. Instead of striving for high-fidelity mapping detail, we consider that the main purpose of a map is to provide environment grounding and context, which we combine with the semantic priors of LLMs to reason about object locations and deploy an active, online approach to navigate to the objects.
Through simulated and real-world experiments we find that our approach tends to have higher retrieval success at shorter path lengths for static objects and by far outperforms prior approaches in cases of dynamic or unmapped object queries. We provide our code and dataset at: 
\href{https://github.com/xiexiexiaoxiexie/osmAG-LLM}{https://github.com/xiexiexiaoxiexie/osmAG-LLM}.
\end{abstract}

\begin{IEEEkeywords}
Semantic Scene Understanding, AI-Enabled Robotics, Vision-Based Navigation
\vspace{-1mm}
\end{IEEEkeywords}

%
\IEEEpeerreviewmaketitle

\vspace{-1mm}
\section{Introduction}
\vspace{-1mm}
%
%
%
%
\IEEEPARstart{N}{avigation} typically requires mapping and planning at multiple scales. Applications such as last mile delivery, inner-city logistics, or assistive navigation systems require city-level street maps to get to the right building, but then also need to navigate inside buildings to go to a specific room or find a requested object. Furthermore, these tasks need understanding about how our world is structured, which items are usually kept in which locations, the usual layouts and functions of different rooms or spaces, as well as other environment-related information.

Multiple prior works \cite{yao2023react,wang2023voyager,park2023generative} found that large language models (LLMs) exhibit very good high-level planning capabilities, especially because they already bring knowledge about many semantic priors in our everyday environments. Robots, however, require not only the general knowledge of LLMs, but also grounding to a specific environment. For instance, while a LLM might suggest searching for scissors in a kitchen, providing the robot with a detailed map of a building allows it to find scissors much quicker in the workshop next door.

Previous works on utilizing LLMs for object navigation, such as 
\cite{qi2020reverie,werby2024hierarchical}, often rely on language-based instructions like ``Bring me the bottom picture that is next to the top of stairs on level one"  or ``find a bean bag in the office on the first floor"  to guide robots. These requests already include a lot of map knowledge in the input to the task,
which may not always be known to the user or the environment might be too large and complex for such explicit guidance.
\begin{figure}[t]
	\centering
	\includegraphics[width=0.95\linewidth]{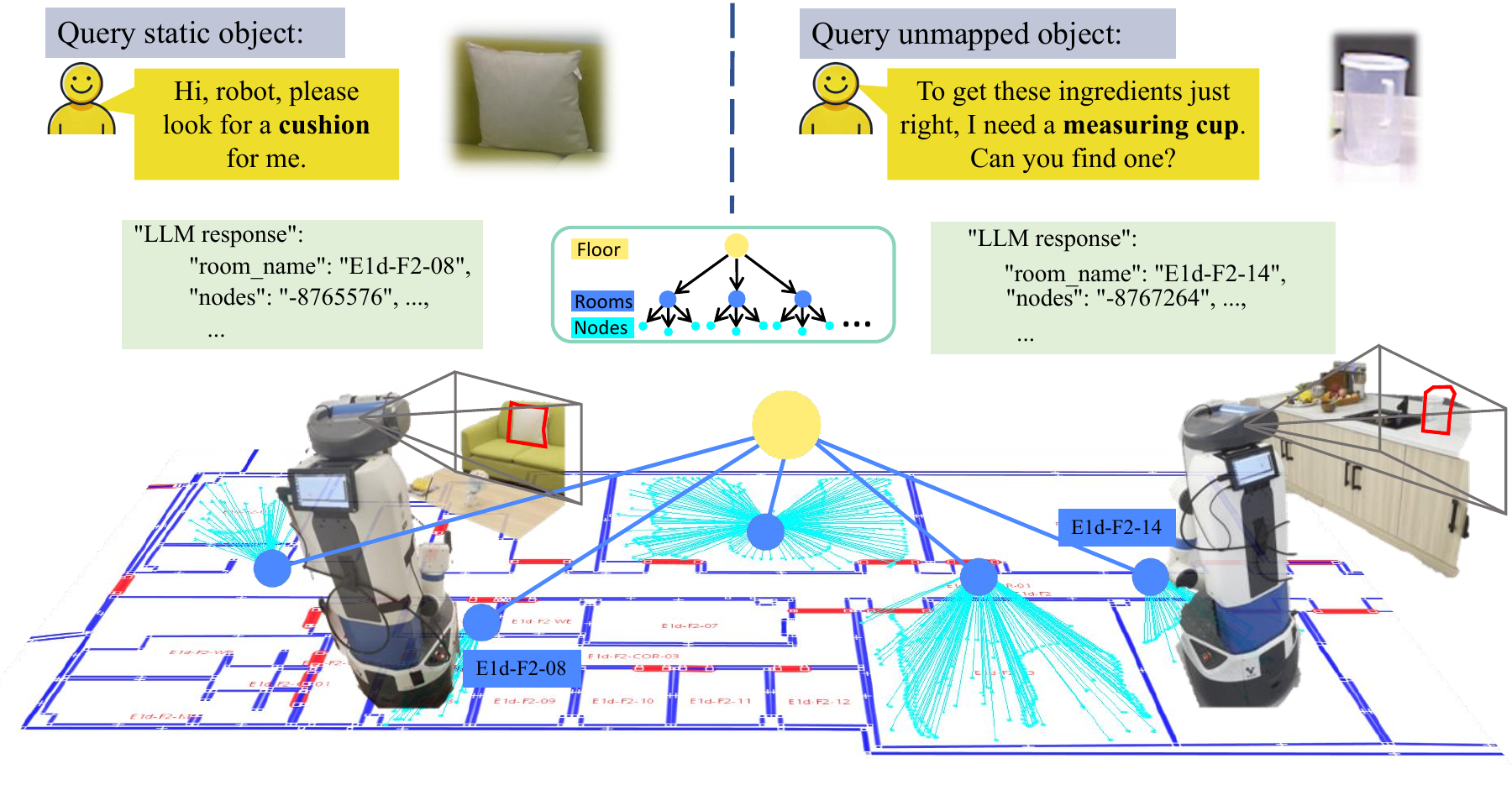}
	\vspace{-4mm}
	\caption{
		The semantic-osmAG employed in our method is a hierarchical, topometric map representation enhanced with textual semantic objects (attached to cyan nodes) and room attributes (attached to rooms). 
		By leveraging this map with LLMs, the robot achieves efficient navigation and objects localization--even for objects absent during initial mapping phase (unmapped objects).
	}
	\label{fig:osmag}
	\vspace{-6mm}
\end{figure}
\begin{figure*}
	\centering
	\includegraphics[width=0.95\linewidth]{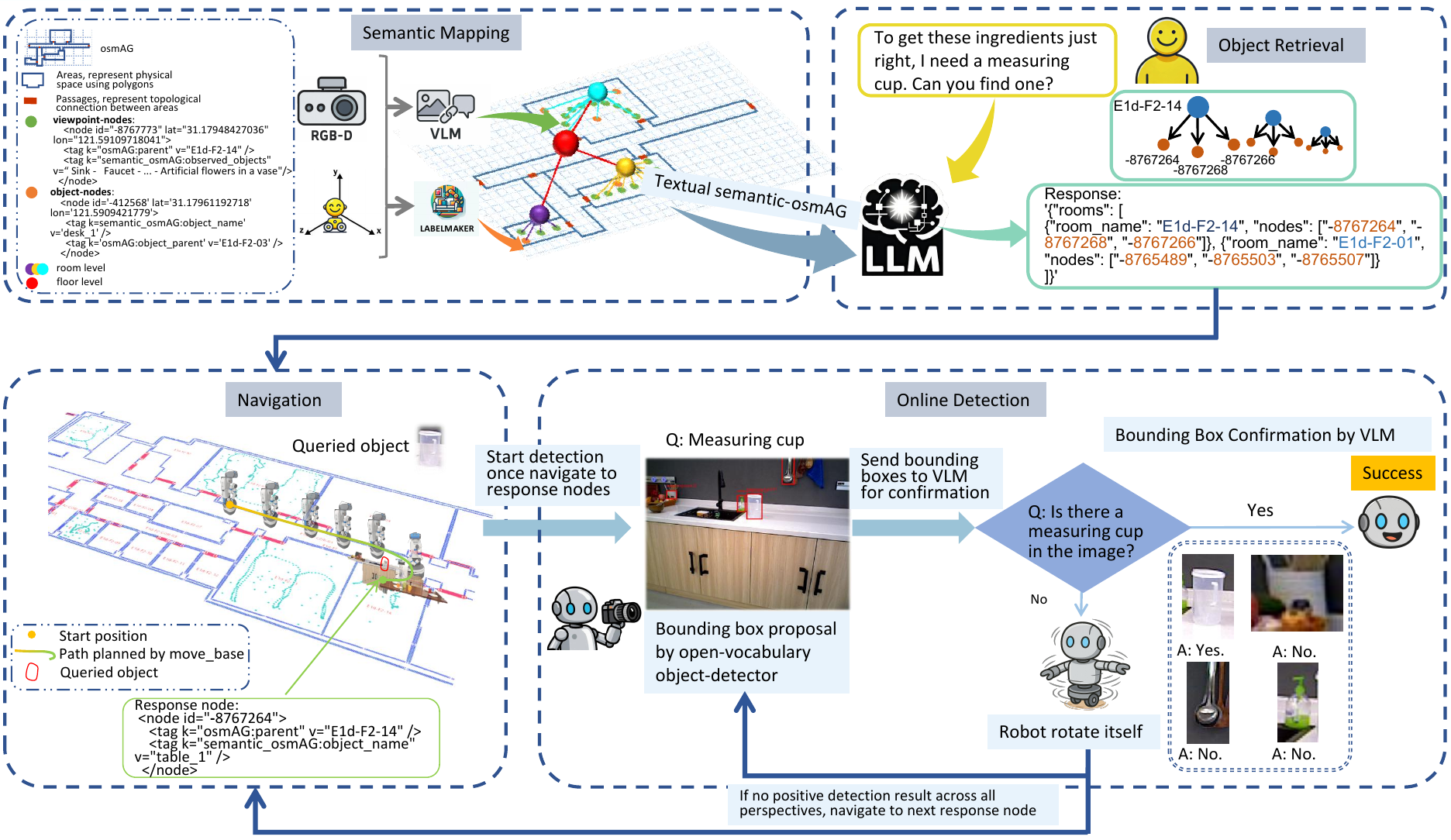}
	\vspace{-2mm}
	\caption{
		An overview of our method. We construct a semantic-osmAG offline by augmenting the basic osmAG with two additional keys: object-nodes (extracted via LabelMaker from RGB-D trajectory data) and viewpoint-nodes (processed by a VLM and placed along the trajectory path).
		When given a human query, the system uses an LLM to generate proposed geometric nodes (response nodes) based on the pre-built textual semantic-osmAG and the query. The robot then navigates to these nodes one by one using ROS move\_base.
		Once at a response node, an open-vocabulary object detector proposes bounding boxes for the queried object, which are then checked by a VLM to verify if the object is actually present.
		If the object isn't found, the robot turns to capture additional perspectives. If the object still isn't detected after checking all views at that node, the robot moves to the next response node and repeats the detection process. {\color{black}(Cartoon robot in Online Detection module generated by ChatGPT-4V; prompt: `robot rotating in place, taking pictures'.)}}
	\label{fig:pipeline}
	\centering
	\vspace{-5mm}
	
\end{figure*}
{\color{black}In this work we also consider less specified queries,}
where a robot is tasked to retrieve a certain type of object, even if it might have moved, there are multiple possible ones in the map, or it has not been mapped at all. As shown in Fig. \ref{fig:osmag}, through a compact, scalable map representations that fits into the context windows of LLMs, we can merge semantic priors and map knowledge to successfully cover all these cases.

In this work, we build upon osmAG (Area Graph
in OpenStreetMap\footnote{\url{https://www.openstreetmap.org/}} textual format)~\cite{feng2023osmag}, a framework that adds room-level mapping to OpenStreetMap, and {\color{black}extends} it to object-level mapping and navigation. Building on osmAG has two key advantages: (i) our system is directly compatible with city-level navigation \cite{hentschel2010autonomous} and can take advantage of the hierarchies defined in OSM; (ii) OSM is a text-based XML format that can directly be parsed by LLMs, without any necessity to collect large-scale data to train encoders and projectors. In particular, we leverage the LabelMaker pipeline~\cite{weder2024labelmaker} as well as VLMs to populate a map with initial object instances and room descriptions. While we prioritize scalability over completeness of the map, we show that through online detection and decision making, we are able to retrieve unmapped objects with close to the same success rate as mapped objects, highlighting the importance of the context a map provides.

In summary, our contributions are:
\begin{itemize}
	\item [$\bullet$] We extend osmAG to object-level semantic mapping, creating a textual semantic map that can be parsed by LLMs.
	\item [$\bullet$] We develop an online object retrieval algorithm that combines the planning capabilities and semantic priors of LLMs with online open-vocabulary detection to retrieve even rare and unusual objects that are not explicitly mapped.
	
	
	\item [$\bullet$] 
	We conduct extensive experiments in simulation and in a real-world campus building, extending the common test cases with object queries that have been moved, as well as querying for unmapped objects.
	
	
\end{itemize} 
To promote reproducibility and further research, we open-source our code, dataset, and related resources.

\section{Related Works}
\subsection{Semantic Maps for LLM-Guided Robotics}

Prior works in semantic maps for robotics fall into two categories: explicit and implicit representations.
Explicit maps use predefined semantic labels (e.g., object classes), offering interpretability but limited flexibility. For example, \cite{gervet2023navigating} construct a 2D semantic map from projecting Mask-RCNN-based instance segmentations. Tag Map \cite{zhang2024tag} represent objects as [id] - [tag] pairs, enabling reasoning with LLMs but lacking geometric structure. 
Guide-LLM \cite{song2024guide} employ a text-based topological map for route planning, assisting visually impaired users through LLM-guided navigation.
Implicit maps leverage vision-language models (e.g., CLIP \cite{radford2021learning}) for open-vocabulary scene understanding. For example, 
OpenScene \cite{peng2023openscene} and OpenMask3D \cite{takmaz2023openmask3d} embed CLIP-aligned features into 3D point clouds or segmented instances, 
ConceptGraphs \cite{gu2023conceptgraphs} and HOV-SG \cite{werby2024hierarchical} further organize instances into 3D scene graphs, with the latter introducing hierarchical open-vocabulary relationships for language-grounded navigation.
MoMa-LLM \cite{honerkamp2024language} utilize a semantic scene graph for long-horizon mobile manipulation, where the robot updates its semantic representation online and uses LLM for task planning.
Unlike these approaches, our method uses semantic-osmAG, a text-based explicit map that contains rich semantic information while retaining hierarchical and topometric structure, enabling traditional robotic localization \cite{xie2023robust} and navigation \cite{xie2024intelligent}.

\subsection{Object Navigation Without Prior Maps}
Prior research in object-goal navigation has explored methods that operate without prior environmental maps. Early approaches, such as SemExp \cite{chaplot2020object}, relied on projective geometry, using Mask R-CNN to extract semantic features from RGB observations and projecting them into 3D voxel maps for reinforcement learning (RL)-based exploration.
Subsequent works address generalization limitations by incorporating vision-language models. For instance, CoWs \cite{gadre2023cows} implement nearest-frontier exploration guided by CLIP's zero-shot recognition. Similarly, VLFM \cite{yokoyama2024vlfm} introduce open-vocabulary frontier scoring, embedding observations via BLIP-2 \cite{li2023blip} and computing semantic similarity to target objects to prioritize exploration paths. GAMap \cite{huang2024gamap} further enhance this approach by integrating object affordances (e.g., ``graspable surfaces") into incrementally constructed maps to guide exploration.

There are also works {\color{black}focused} on improving spatial reasoning using LLMs. For example, ESC \cite{zhou2023esc} employ LLMs as rule-based frontier selectors, leveraging commonsense reasoning to infer spatial relationships between target objects and common objects or room layouts. 
Likewise, \cite{arjun2024cognitive} utilize LLMs' commonsense knowledge to predict object locations.

However, these methods do not rely on pre-built maps, which can lead to challenges in scenarios such as locating a toilet behind a closed bathroom door or navigating long corridors where doorways are not immediately visible. In contrast, our method leverages pre-built maps to address these limitations.

\subsection{Object Navigation with Prior Maps}
Prior research has also explored object-goal navigation using pre-built maps.
ConceptGraphs \cite{gu2023conceptgraphs} generate 3D scene graphs by projecting class-agnostic segments into metric maps, embedding them with CLIP, and inferring spatial relationships via LLMs.
Other graph-based methods like SEEK \cite{Ginting2024Seek} utilize dynamic scene graphs (DSGs) derived from environmental blueprints, training a neural network to estimate the probability of finding the target object across spatial elements in the graph.
HOV-SG \cite{werby2024hierarchical} construct a hierarchical open-vocabulary 3D scene graph that enables hierarchical query and object navigation.
Unlike prior methods that rely on dense representations, our approach: 
1. avoids detailed 3D environment mapping to ensure long-term usability, 
2. uses powerful VLMs to summarize environmental objects, 
3. maintains a lightweight textual map with hierarchy and topology that preserves key semantic relationships for LLM-based reasoning.

\section{Approach}
This work tackles the challenge of map-guided object navigation by combining semantic maps (semantic-osmAG, based on osmAG \cite{feng2023osmag} as shown in Fig. \ref{fig:osmag}) with large language models. As illustrated in Fig. \ref{fig:pipeline}, our approach begins by augmenting the basic osmAG map using RGB-D trajectories to build a semantic-rich map representation for subsequent tasks (Section \ref{sec:mapping}). Next, given a human query, an LLM is leveraged to infer probable map nodes corresponding to the target object (Section \ref{sec:object_retrieval}). Finally, the robot navigates toward the nodes (Section \ref{sec:navigation}), with navigation further refined through real-time online object detection (Section \ref{sec:online_detection}).

\subsection{Semantic Environment Mapping}
\label{sec:mapping}
\subsubsection{Map Representation}
While traditional maps (e.g., occupancy grids, point clouds) excel at spatial representation, their raw numeric form offers little semantic structure and is therefore difficult for a large language model to understand. We instead employ semantic-osmAG--an indoor, semantic extension of the osmAG map--because it provides four decisive advantages:

\begin{enumerate}
	\item Human-LLM Interpretable Format: osmAG uses OSM standards, with editable key-value properties for nodes and rooms, naturally readable by both humans and LLMs without extra conversion.
	
	\item Compact Representation:
	As Section \ref{sec:osmag_mapping} describes, osmAG encodes areas (rooms) as simple polygons, objects as single nodes with pure text format, enabling large environment deployment and sharing through OSM server.
	
	\item 
	Hierarchical Object Retrieval: 
	The osmAG map uses parent tags to represent which objects are contained in which rooms, enabling LLMs to reason about probable locations. 
	For example, when searching for a hot air gun, the LLM recognizes it more likely belongs in a robotics-equipped lab than a kitchen, and directs the search accordingly.
	Once the target room is selected, the same hierarchy is used to identify specific nodes within the room.
	This closely mimics human search behavior by allowing the system to narrow the search from a broad room-level down to precise locations, rather than relying on memorizing the entire environment's point cloud.
	\item Minimal Update Required:
	Through semantic filtering, we can exclusively utilize permanent infrastructure (e.g., walls, doors) for osmAG based robot localization \cite{xie2023robust} and navigation \cite{xie2024intelligent} and avoid saving pixel level object representations. Therefore, semantic-osmAG naturally adapts to dynamic environments and requires minimal updates.
	
\end{enumerate}
\subsubsection{osmAG}
\label{sec:osmag_mapping}
osmAG\cite{feng2023osmag} is a hierarchical, topometric semantic map representation based on the Area Graph concept \cite{hou2019area}. As illustrated in Fig. \ref{fig:osmag} and used in our experiments (Fig. \ref{fig:experiment}), osmAG models physical spaces (e.g., rooms, corridors) as polygon-based areas and uses passages (e.g., doors) to represent connections between areas. Moreover, osmAG employs a `parent' tag to explicitly represent hierarchical relationships between spaces (e.g., rooms within a building or floor).
osmAG can be constructed from various sources, including 3D point clouds \cite{he2021hierarchical}, occupancy grid maps \cite{hou2019area}, or CAD files \cite{feng2023osmag,zhang2025generation}.
Prior works have established osmAG's utility for core robotic tasks: robust LiDAR-based localization \cite{xie2023robust} and intelligent navigation \cite{xie2024intelligent}. 
While osmAG originally lacked the granularity to locate specific objects (e.g., a hot air gun), Section \ref{sec:semantic_enrichment} presents our semantic enrichment method to address this limitation.

\begin{figure}[t]
	\centering
	\includegraphics[width=0.99\linewidth]{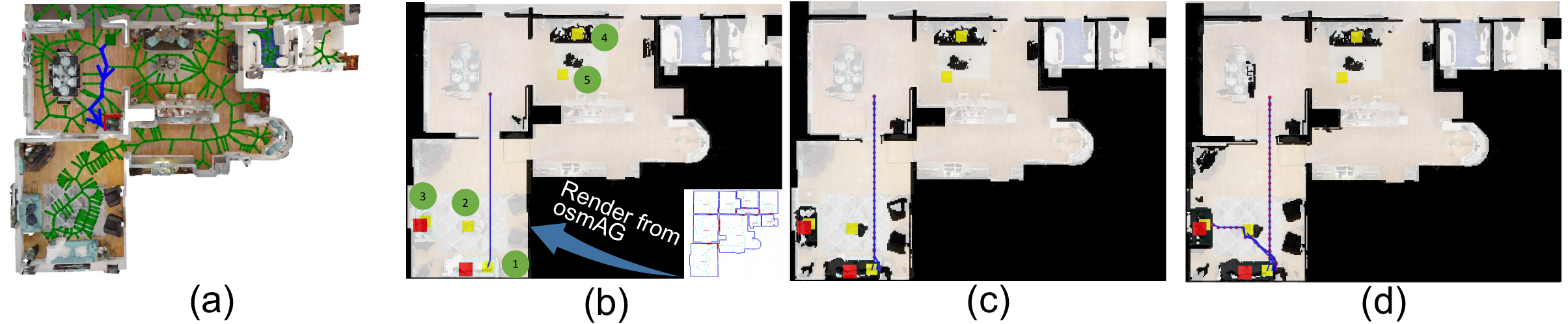}
	\vspace{-4mm}
	\caption{Figure (a) shows a pre-built navigation graph from \cite{werby2024hierarchical} for reference, while (b)-(d) show our navigation process. With query `couch in the living room': 
		(b) initial setup with an osmAG-rendered occupancy map (walls/doors only), where red rectangles mark ground truth and yellow rectangles show response nodes (green circles highlighting their sequence).
		(c)-(d) The robot progressively perceives the environment, navigating nodes (1$\to$3) and replanning upon collisions.
		Our navigation strategy produces more direct paths compared to pre-built navigation graph in (a), and through progressive environmental perception, achieves long-term operation with minimal map updates required.}
	\label{fig:navigation}
	\vspace{-4mm}
	
\end{figure}
\subsubsection{Semantic Enrichment}
\label{sec:semantic_enrichment}
To integrate semantic information into osmAG, we developed a pipeline for generating semantic nodes from RGB-D data. 
We generate the object-nodes by first utilizing LabelMaker\cite{weder2024labelmaker} to process RGB-D trajectories through ensemble voting across state-of-the-art models to produce point clouds for each object category. Next, we use Mask3D's \cite{schult2023mask3d} instance masks to isolate individual instances from category point clouds. We calculate each instance's centroid and embed it as a `node' (with longitude/latitude coordinates) within the osmAG structure with semantic\_osmAG:object\_name as tag's key as illustrated in Fig. \ref{fig:pipeline}.
However, LabelMaker inherits the categorical limitations of its underlying models, failing to recognize objects outside standard datasets like ScanNet200 \cite{rozenberszki2022language}, which limits its ability to process unrestricted natural language queries about objects (e.g., robot dog).
To address this, we augment the system with ChatGPT's \cite{achiam2023gpt} visual query understanding capability. By prompting with `What objects are in this image?' at capture locations, we generate viewpoint-nodes at the image's captured location with tag key as semantic\_osmAG:observed\_object and value as the response of VLM as illustrated in Fig. \ref{fig:pipeline}. 
This hybrid approach -- combining geometric instance extraction and open-vocabulary visual language response -- generates part of the nodes with exact geometric location and gives part of the nodes open-vocabulary capability from a more intelligent model, overcoming categorical rigidity inherent in LabelMaker alone.

Beyond embedding geometric object nodes, we generate semantic room descriptions by prompting a VLM with `Describe the image.' for all images captured in a room. The resulting descriptions are then summarized by an LLM and attached directly to the corresponding area elements in the semantic-osmAG map.
\vspace{-2mm}
\subsection{Object Retrieval}
\vspace{-1mm}
\label{sec:object_retrieval}
When processing a human query, our method generates a structured LLM prompt that: (1) explains semantic-osmAG's representation, (2) defines the object retrieval task objectives, and (3) provides a simplified semantic-osmAG containing only semantic nodes and hierarchical relationships (excluding coordinates and polygons).

To optimize search efficiency and align with practical navigation patterns, we specify to the LLM to generate results using a JSON format where response nodes {\color{black}(which can be either viewpoint or object nodes)} are organized by room (max 3 nodes per room and max 3 rooms total), ordered by decreasing predicted likelihood as shown in Fig. \ref{fig:pipeline}.
This room-centric approach mimics human search behavior - checking one room completely before moving to the next, rather than jumping between individual nodes based solely on probability.

\vspace{-2mm}
\subsection{Navigation}
\vspace{-1mm}
\label{sec:navigation}

As illustrated in Fig. \ref{fig:navigation}, unlike methods like HOV-SG \cite{werby2024hierarchical} using pre-built navigation graphs, we dynamically navigate via online perception from rendered osmAG. We first convert osmAG to an occupancy grid (area polygons as occupied, passages as free). The system then sequentially visits response nodes from LLM: (1) setting navigation goals from sorted nodes, (2) planning paths via A* on the occupancy grid, and (3) continuously {\color{black}updating} the rendered occupancy grid map using real-time depth data and {\color{black}triggering} path replanning when collisions are detected in the updated map. By building the grid only from permanent structures and adapting online, we avoid needing map updates when environments change.
\begin{table*}[h]
	\caption{Object Retrieval \& Goal Navigation from Language Queries on HM3DSEM}
	\vspace{-2mm}        
	\label{table:habitat}
	\renewcommand{\arraystretch}{0.8}
	\centering
	\begin{tabular}{lllccccccccc}
		\toprule
		\multirow{2}{*}{Query Type} & \multirow{2}{*}{Method} & \multirow{2}{*}{R-RSR} & \multicolumn{3}{c}{O-RSR$_{top5}$} & \multicolumn{3}{c}{O-RSR$_{top1}$} & \multirow{2}{*}{AMD [m]} & \multirow{2}{*}{DIR } & \multirow{2}{*}{APL [m]} \\
		\cmidrule(lr){4-6} \cmidrule(lr){7-9}
		& & & @1m & @2m & @3m & @1m & @2m & @3m & & & \\
		\midrule
		\multirow{2}{*}{(o,r,f)} & HOV-SG & 0.27 & \textbf{0.31} & 0.39 & 0.44 &\textbf{ 0.20} & 0.28 & 0.33 & 8.85 & 0.05 & 20.15 \\
		& our method & \textbf{0.66} & 0.25 & \textbf{0.59} & \textbf{0.70} & 0.16 & \textbf{0.37} & \textbf{0.54} & \textbf{4.65} & \textbf{0.27} & \textbf{12.32} \\
		\cmidrule(lr){1-12}
		
		\multirow{2}{*}{(o,r)} & HOV-SG & 0.32 & \textbf{0.35} & 0.44 & 0.49 & \textbf{0.20} & 0.28 & 0.33 & 7.87 & 0.06 & 16.81 \\
		& our method & \textbf{0.67} & 0.28 & \textbf{0.58} & \textbf{0.68} & 0.18 & \textbf{0.37} & \textbf{0.55} & \textbf{5.27} & \textbf{0.25} & \textbf{11.54} \\
		\cmidrule(lr){1-12}

		\multirow{2}{*}{(o)} & HOV-SG & 0.65 & \textbf{0.76} & 0.82 & 0.88 & \textbf{0.52} & \textbf{0.64} & \textbf{0.70} & \textbf{3.10} & 0.05 & 22.44 \\
		& our method & \textbf{0.83} & 0.47 & \textbf{0.83} & \textbf{0.90} & 0.28 & 0.50 & 0.69 & 3.37 & \textbf{ 0.31} & \textbf{11.36} \\

		\bottomrule
	\end{tabular}
	\vspace{-6mm}
\end{table*}
\vspace{-2mm}
\subsection{On-line Object Detection}
\vspace{-1mm}
\label{sec:online_detection}
To ensure robust adaptation to environmental changes and verify object presence at response nodes, our system performs real-time detection upon arrival. As shown in Fig. \ref{fig:pipeline}, when the robot arrives at a response node, it captures images and executes a two-stage verification process.
We run open-vocabulary object-detectors (Yolo-world \cite{cheng2024yolo} or DINO-X \cite{ren2024dino}) on the captured images to generate a list of bounding boxes ranked by confidence. Each box is cropped and validated by a VLM (StepFun \cite{stepfun_step1o_turbo_vision}) using the prompt ``Is there a [object] here?", with the highest-confidence positive detection returned as the final result.
We find that this two-step approach is more reliable with long-tail object queries where the detectors would otherwise return too many false positives.
If no object is detected, the robot rotates its body or head to capture additional viewpoints, repeating the verification process until either (1) the target object is successfully identified, or (2) all potential viewpoints at the current node have been exhausted without positive confirmation, at which point it proceeds to the next response node.

\section{Experiments}
We evaluated our approach on eight scenes from the HM3D-SEM dataset \cite{yadav2023habitat} (consistent with our baseline) as well as on a real-world dataset collected in our campus building.
In Section \ref{sec:baseline}, we justify our selection of HOV-SG\cite{werby2024hierarchical} {\color{black} and ConceptGraph\cite{gu2023conceptgraphs}} as {\color{black}baselines}. Section \ref{sec:evaluate_metrics} details our evaluation metrics, followed by Habitat simulation experiments using the HM3D-SEM dataset in Section \ref{sec:exper_habitat}, and real-world experiments in Section \ref{section:exper_real}.

\subsection{Baseline}
\label{sec:baseline}
Current object-aware mapping approaches suffer from key limitations: Hydra \cite{hughes2022hydra} and SEEK \cite{Ginting2024Seek} lack open-vocabulary support, restricting their utility for real-world queries (e.g., ``hot air gun").
VLMaps \cite{huang2023visual} fails to incorporate hierarchical representations, potentially compromising scalability in large environments.

We select HOV-SG as our baseline because it shares fundamental design principles with our approach. Both methods: (1) rely on pre-built semantic maps for environment representation, (2) utilize natural language object queries as their primary input  without describing how to get to the location, (3) have the same input mapping modalities (RGB-Depth-Pose data pairs), and (4) employ a hierarchical map structure. 

To ensure a fair comparison, we directly incorporate HOV-SG's query prompt formulation in our experiments using the HM3D-SEM dataset.
{\color{black}In addition to HOV-SG, we include ConceptGraph \cite{gu2023conceptgraphs} as an additional baseline in our real-world experiments to provide a more comprehensive evaluation. On the HM3D-SEM dataset, we refer to the evaluation in~\cite{werby2024hierarchical} from which we identify HOV-SG as the stronger baseline for our experiments.}




\textcolor{black}{We note that} the original HOV-SG's output point clouds are not room-aware, causing the agent to frequently jump between rooms (e.g., leaving Room 1 and later returning to it), thereby unnecessarily increasing the path length. To mitigate this, we improve the baseline by re-sorting the top 5 HOV-SG predictions to prioritize visiting all nodes in one room before moving to another.


\subsection{Metrics}
\label{sec:evaluate_metrics}
We employ the following metrics to evaluate performance across both simulated and real-world environments:
\begin{itemize}
	\item \textbf{Room Retrieval Success Rate (R-RSR)}: Percentage of queries where the top prediction's room contains at least one ground truth instance of the target object.

	\item \textbf{Object Retrieval Success Rate (O-RSR$_{top-n}$@k)}: Percentage of queries where at least one top-$n$ prediction ($n\in{1,5}$) is within $k$ meters (Euclidean distance) of ground truth. We test $k\in{1,2,3}$m to accommodate positional variance between our 2D image-derived nodes and HOV-SG's 3D point clouds.
	
	\item  \textbf{Average Minimum Distance (AMD)}: 
	Computes the mean distance error of the closest prediction among the top-5 results to evaluate object localization precision.
	

	
	
	
	\item \textbf{Average Path Length (APL)}: Mean navigation path length that the robot drove (in meters) for queries where both our method and baseline successfully retrieve the target (within 1m). Measures navigation efficiency.

	\item \textbf{Detection Improvement Rate (DIR)}:  
	Percentage of queries where initially failed retrievals (where all top-5 predicted nodes exceed 1m from ground truth) that are successfully recovered through online detection at response nodes.
	
	\item \textbf{Map size}: Compares the representation size of our method and HOV-SG.
\end{itemize}

\subsection{HM3D-SEM Dateset Experiment}
\label{sec:exper_habitat}

\subsubsection{Setup}
We evaluate our approach on 8 diverse HM3D-SEM scenes (00824, 00829, 00843, 00861, 00862, 00873, 00877, 00890) which span multiple rooms and floors. 
Each scene is tested with 5 randomly chosen starting positions, with the Habitat-simulated robot equipped with an RGB-D camera for perception.
The base osmAG map is created manually using JOSM (Java OpenStreetMap Editor)\footnote{\url{https://josm.openstreetmap.de/}} tool. We then enhance it semantically using LabelMaker and ChatGPT-4V, processing the same RGB-D trajectories as HOV-SG to generate the final representation shown in Fig. \ref{fig:navigation} (b).
Navigation follows our defined approach (Section \ref{sec:navigation}, Fig. \ref{fig:navigation}). Online detection is implemented through a two-stage process: YOLO-World generates initial bounding box proposals, followed by StepFun's confirmation reasoning.
We generate queries using the same approach as HOV-SG with three granularity levels for 20 common objects: 1. object (o): ``pillow";  2. object + room (o,r): ``pillow in the living room"; 3. object + room + floor (o,r,f): ``pillow in the living room on floor 0".

\subsubsection{Results}
\label{sec:results}

Table \ref{table:habitat} presents comprehensive evaluation results across three query granularities, our method demonstrates significant improvements in key metrics over the HOV-SG baseline:

\begin{itemize}
	\item Superior room and object retrieval: Notably, we achieve 144\% higher Room Retrieval Success Rate (R-RSR) for complex (o,r,f) queries (0.66 vs. 0.27) and maintain better performance at practical operating distances (O-RSR@2m/3m), despite HOV-SG's advantage in precise close-range (@1m) object retrieval due to its direct point-cloud outputs. The AMD metric confirms our predicted nodes are closer to ground truth positions in (o,r) and (o,r,f) query types.
	
	\item Hierarchical advantage: Both methods exhibit performance degradation as query complexity increases, but our approach demonstrates significantly more graceful degradation compared to HOV-SG. 
	
	\item Effective online detection compensation: The online detection system effectively compensates for retrieval imperfections, as evidenced by substantially higher DIR (Detection-Improvement Rate).
	
	\item Efficient navigation: The Average Path Length (APL) metric shows that our method reduces traversal distance by 31-49\% across all conditions. This confirms that navigation using osmAG is more effective than navigation using the pre-build navigation graph of the baseline (illustrated in Fig. \ref{fig:navigation}).
	\item Significantly smaller storage requirements: As shown in Table \ref{table:size}, our method requires substantially less storage space than the baseline  {\color{black}(while HOV-SG's representation scales with computation time rather than context size, this map-size comparison highlights the compactness and storage efficiency of our method)}.
	\begin{table}[ht]
		\vspace{-3mm}
		\caption{Representation Size Comparison (MB)}
		\vspace{-2mm}
		\renewcommand{\arraystretch}{0.8}
		\centering
		\begin{tabular}{
				>{\centering\arraybackslash}p{1.5cm}  
				>{\centering\arraybackslash}p{3cm}    
				>{\centering\arraybackslash}p{2cm}    
			}
			\toprule
			& HM3D-SEM (8 scenes) & Real Dataset \\ 
			\midrule
			HOV-SG   & 1493   & 129   \\ 
			Our method   & 3.2   & 0.62   \\ 
			\bottomrule
			
		\end{tabular}
		\label{table:size}
		\vspace{-7mm}
	\end{table}
	
\end{itemize}


\begin{figure}[t]
	\includegraphics[width=0.99\linewidth]{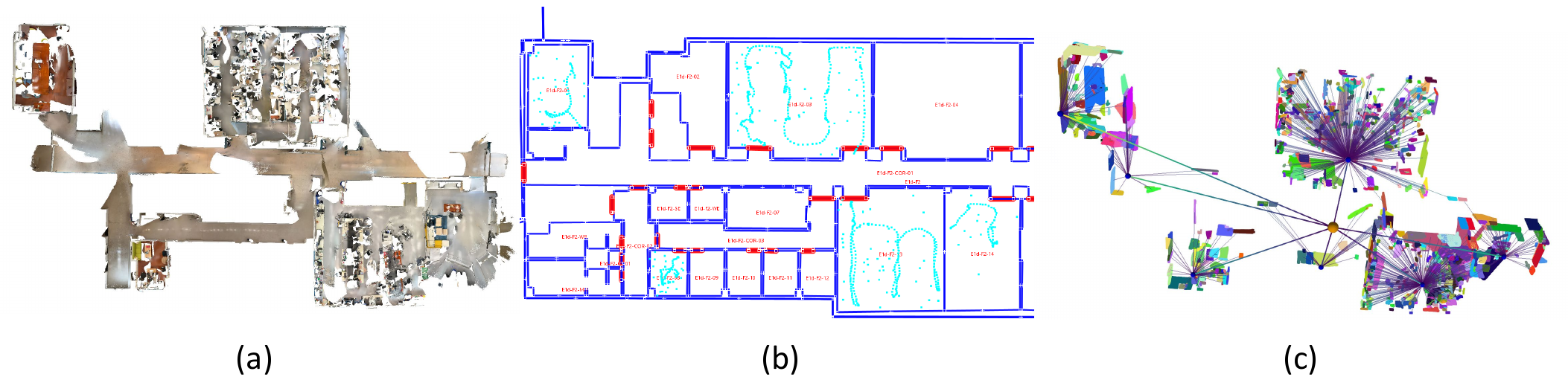}
	\vspace{-2mm}
	\caption{Real-world experimental environment consisting of: a conference room (upper left), a student office (upper right), a professor's office (lower left), a robotics lab (lower middle), and a relaxation lounge (lower right). (a) Data collected using an Apple scanner with the ``3D Scanner App"; (b) the semantic-osmAG map; (c) the HOV-SG generated from the collected data. }
	\label{fig:experiment}
	\centering
	\vspace{-5mm}

\end{figure}
\subsection{Real-World Environment Experiment}
\label{section:exper_real}
\subsubsection{Setup}
\label{sec:real_exper_setup}
We conducted experiments in a large and cluttered campus building spanning multiple functional areas: a conference room (46 $m^2$), a robotics lab (120 $m^2$), a student office (147 $m^2$), a professor's office (15 $m^2$), and a relaxation lounge (82 $m^2$) (Fig. \ref{fig:experiment}). 
To scale our approach to larger environments
{\color{black}, we further propose a two-stage query variant. The first stage uses only the room-level descriptions for coarse room selection, while the second stage provides the detailed map of the chosen room to select specific target nodes.}
{\color{black}To analyze the trade-off in map complexity, we also investigated the impact of viewpoint node density--examining how performance varies when too few nodes may provide insufficient context versus when too many may confuse the LLM.}

For scanning, we exclusively used an Apple scanner with the `3D Scanner App'\footnote{\url{https://3dscannerapp.com/}}.
Unlike the high-fidelity HM3D-SEM dataset, our environment was intentionally mapped with lightweight scanning to evaluate practical deployment feasibility, resulting in less detailed spatial representations (Fig. \ref{fig:experiment}a).
The robotic platform consisted of a Fetch mobile manipulator equipped with: A base-mounted 2D LiDAR for localization and navigation (using ROS AMCL and move\_base packages), a head pan link capable of 180-degree rotation to capture diverse viewpoints, and a head-mounted RealSense RGB-D camera for visual perception.
As the student office contains steps that are not traversable with Fetch, we implemented two methodological adaptations: 1) Manual camera positioning at designated response nodes for object detection. 2) Use of planned (rather than executed) global paths from move\_base for Average Path Length (APL) metrics.
These adaptations were applied to both our method and HOV-SG, maintaining comparative fairness.     
For both methods, we further simplify testing by skipping physical navigation when the query object was definitively absent from a room, again using move\_base's global path as Path Length. 

\begin{figure}[t]
	\centering
	\includegraphics[width=0.99\linewidth]{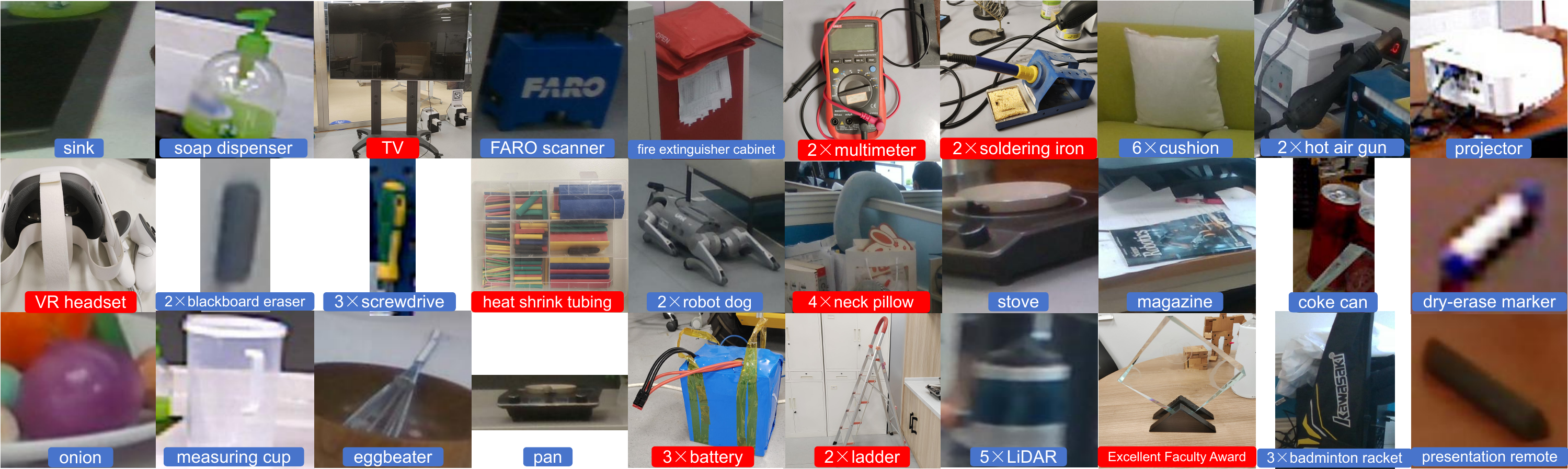}
	
	\caption{
		Experimental objects used for real-world evaluation: The {\color{black}first row shows} static objects, {\color{black}the second row contains} relocated objects, and the final {\color{black}row displays} unmapped objects absent from the map. Successful detections are shown with automatically generated bounding boxes during experiments, while text overlays indicate object names and instance counts in our environment (blue = success, red = failure).
	}
	
	\label{fig:objects}
	\vspace{-5mm}
	
\end{figure}

\subsubsection{Queries}
\label{sec:queries}
To evaluate both methods under realistic conditions, we conducted 30 experiments divided into three distinct categories (shown in Fig. \ref{fig:objects}):
Static Objects (SO, e.g., sink/TV) remaining approximately at mapped positions; Relocated Objects (RO, e.g., screwdriver/robot dog) moved from initial locations; and Unmapped Objects (UO, e.g., presentation remote/measuring cup) absent during mapping. Each category (SO, RO, UO) contained six unique single-location objects and four multi-instance objects, totaling 11, 12, and 13 objects per category respectively.
\begin{figure}[b]
	\vspace{-5mm}
	
	\centering
	\includegraphics[width=0.99\linewidth]{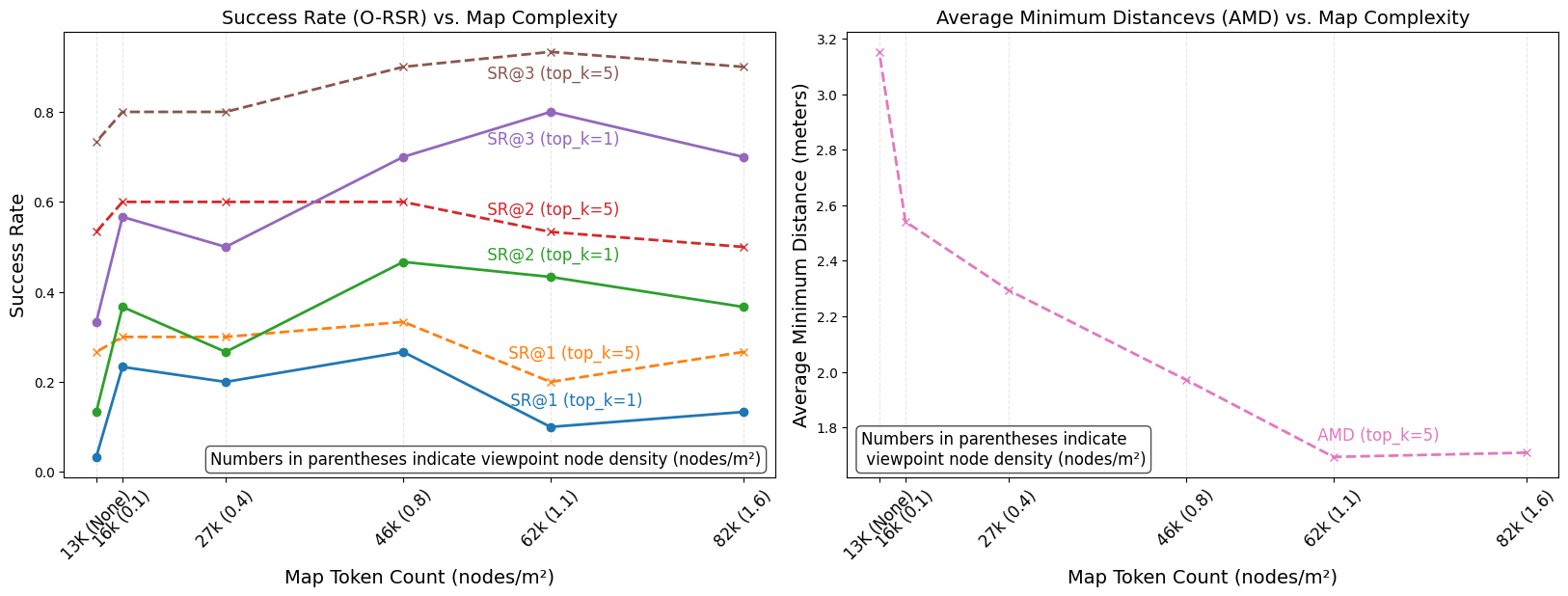}
	\vspace{-4mm}
	\caption{{\color{black}
			Object retrieval performance versus map complexity, represented by map token count and its corresponding viewpoint node density (parenthetical). (Left) Object Retrieval Success Rate (O-RSR) and (Right) Average Minimum Distance (AMD).
		}
	}
	\label{fig:viewpoint_nodes}
	\vspace{-5mm}
\end{figure}

\begin{table*}[h]
	\caption{Object Retrieval \& Goal Navigation from Language Queries on Real Data}
	\vspace{-2mm}        
	\label{table:real_exp}
	\renewcommand{\arraystretch}{0.8}
	\centering
	\begin{tabular}{lllcccccccccc}
		\toprule
		\multirow{2}{*}{Object Type} & \multirow{2}{*}{Method} & \multirow{2}{*}{R-RSR} & \multicolumn{3}{c}{O-RSR$_{top5}$} & \multicolumn{3}{c}{O-RSR$_{top1}$} & \multirow{2}{*}{AMD [m]} & \multirow{2}{*}{DIR} & \multirow{2}{*}{APL [m]} \\
		\cmidrule(lr){4-6} \cmidrule(lr){7-9}
		& & & @1m & @2m & @3m & @1m & @2m & @3m & & & \\
		\midrule
		\multirow{3}{*}{static (SO)} &\color{black}ConceptGraph & 0.50  & \textbf{0.50} & 0.50 & 0.60 & \textbf{0.20} & 0.30 & 0.40 & 3.52 (3.52) & / & / \\
		
		& HOV-SG & 0.40 & 0.20 & \textbf{0.70} & \textbf{0.80 }& 0.00 & 0.30 & 0.30 & 1.84 (1.84) & 0.40 (0.40) & 47.00 \\
		
		& our method & \textbf{1.00} & 0.30 & 0.50 & \textbf{0.80} & \textbf{0.20} & \textbf{0.40} & \textbf{0.80} & \textbf{1.78 (1.78)} & 0.40 (0.40) & \textbf{18.13} \\
		\cmidrule(lr){1-12}

		\multirow{3}{*}{relocated (RO)} & \color{black}ConceptGraph & 0.50 & 0.20 & 0.30 & 0.50 & 0.00 & 0.10 & 0.20 & 5.51 (5.86) & / & / \\
		
		& HOV-SG & 0.40 & \textbf{0.30} & 0.50 & 0.50 & 0.00 & 0.20 & 0.20 & 5.36 (5.79) & 0.00 (0.00) & 22.05 \\
		
		& our method & \textbf{1.00}  & \textbf{0.30 }& \textbf{0.70} & \textbf{0.90} & \textbf{0.20} & \textbf{0.60} & \textbf{0.80} & \textbf{1.63 (1.72)} & \textbf{0.70 (0.78)} & \textbf{20.43} \\
		\cmidrule(lr){1-12}

		\multirow{3}{*}{unmapped (UO)} & \color{black}ConceptGraph & 0.40 & \textbf{0.10 }& 0.20 & 0.30 & 0.00 & \textbf{0.10} & 0.20 & 8.35 (8.83) & / & / \\
		
		& HOV-SG & 0.20 & 0.00 & \textbf{0.30} & 0.50 & 0.00 & 0.00 & 0.20 & 7.42 (7.63) & 0.30 (0.33)& 52.03 \\
		
		& our method & \textbf{1.00} & 0.00 & \textbf{0.30} & \textbf{0.80} & 0.00 & \textbf{0.10} & \textbf{0.60} & \textbf{2.25 (2.33)} & \textbf{0.80 (0.78)} & \textbf{30.68} \\
		
		\bottomrule
	\end{tabular}
	
	\begin{tablenotes}
		\item[a] {\color{black}Values in parentheses indicate: results after excluding inaccessible rooms (Section \ref{sec:real_exper_setup}), maintaining the same trends as overall results.}
	\end{tablenotes}
\vspace{-5mm}
\end{table*}

\subsubsection{Results}
Our real-world experiments demonstrate consistent advantages over baseline (Table \ref{table:real_exp}):


\begin{itemize}
	\item Superior room and object retrieval: 
	Our method achieves 100\% room retrieval, surpasses {\color{black}baselines} in O-RSR metrics as distance increases, and maintains lower AMD (nodes closer to ground truth).
	\item Effective online detection compensation: 
	Our higher DIR provides strong evidence that our online detection approach effectively compensates for sparse mapping requirements while eliminating the need for precise 3D point cloud data saved in the map. 
	
	\item Efficient navigation: The Average Path Length (APL) metric shows our method reduces path length across all conditions.
	\item Environmental robustness: Furthermore, our method significantly outperforms {\color{black}baselines} on relocated and unmapped objects, highlighting its robustness to environmental variability.
	
	{\color{black}\item Two-stage query performance: The results in Table \ref{table:ablation} reveal a surprising effectiveness of this approach for static objects. For relocated or unmapped objects, the two-stage approach is less successful than when providing context of all the room immediately, but still on par or better than the baselines. Overall these indicate promising scalability of osmAG-LLM for very large environments.}
	
	{\color{black}\item Viewpoint node density: Our experiment reveals a non-linear relationship between viewpoint node density and performance (Fig. \ref{fig:viewpoint_nodes}). A small number of nodes yields substantial performance gains, offering high reward.
		However, beyond a certain threshold, additional nodes provide diminishing returns and may even slightly degrade performance due to potential information overload. Overall the density does not seem to require critical tuning as long as some viewpoint nodes for every room are provided.}
	
\end{itemize}

\subsubsection{Ablation Studies}
\label{sec:ablation_study}
To evaluate the contributions of the two key components in our map--object nodes and viewpoint nodes--we conduct an ablation study using real-world data (Table \ref{table:ablation}). First, we remove all object nodes generated by LabelMaker. This degrades RSR and AMD performance, since the system loses precise object location information. Next, we eliminate all viewpoint nodes generated by the VLM, which leads to rapid performance degradation, since our language queries contain many objects outside our detection model's fixed category set. Our results demonstrate that both node types are essential, as removing either component leads to substantially worse performance compared to the complete method. 
\begin{table}[t]
	\caption{{\color{black}Comparison of Method Variants on Real Data}}    
	\vspace{-2mm}    
	\label{table:ablation}
	
	\centering

	\scalebox{0.85}
	{
		\setlength{\tabcolsep}{2pt} 
		\renewcommand{\arraystretch}{0.9} 
		\begin{tabular}{@{}lcc*{6}{c}c@{}}
			\toprule
			\multirow{2}{*}{Object} & \multirow{2}{*}{Method} & \multirow{2}{*}{R-RSR} & \multicolumn{3}{c}{O-RSR$_{top5}$} & \multicolumn{3}{c}{O-RSR$_{top1}$} & \multirow{2}{*}{AMD [m]} \\
			\cmidrule(r){4-6} \cmidrule(l){7-9}
			Type& & & @1m & @2m & @3m & @1m & @2m & @3m & \\
			\midrule
			\multirow{4}{*}{SO} & Ours & \textbf{1.0} & 0.3 & 0.5 & 0.8 & 0.2 & 0.4 & \textbf{0.8} & 1.78 \\
			
			
			&\color{black}Ours-2stages& \textbf{1.0} & \textbf{0.4} & \textbf{0.8} & \textbf{1.0} & \textbf{0.3} & \textbf{0.5} & \textbf{0.8} & \textbf{1.30} \\
			
			& w/o ON & \textbf{1.0} & 0.3 & 0.5 & 0.7 & \textbf{0.3} & 0.4 & 0.6 & 2.14 \\
			& w/o VN & 0.9 & 0.2 & 0.5 & 0.6 & 0.1 & 0.2 & 0.5 & 3.77 \\
			\cmidrule(r){1-10}
			
			\multirow{4}{*}{RO} & Ours & \textbf{1.0} &\textbf{0.3} & \textbf{0.7} & \textbf{0.9} & 0.2 & \textbf{0.6} & \textbf{0.8} & \textbf{1.63} \\
			
			
			& \color{black}Ours-2stages& 0.7 & \textbf{0.3} & 0.5 & 0.7 & 0.2 & 0.4 & 0.6 & 7.44 \\
			& w/o ON & 0.9 & \textbf{0.3} & \textbf{0.7} & 0.8 & \textbf{0.3} & 0.5 & 0.6 & 2.00 \\
			& w/o VN & 0.7 & 0.2 & 0.2 & 0.5 & 0.0 & 0.1 & 0.2 & 4.18 \\
			\cmidrule(r){1-10}
			
			\multirow{4}{*}{UO} & Ours & \textbf{1.0}& 0.0 & \textbf{0.3} & \textbf{0.8} & 0.0 & \textbf{0.1} & \textbf{0.6} & \textbf{2.25} \\
			

			& \color{black}Ours-2stages& 0.9 & 0.0 & 0.2 & 0.7 & 0.0 & 
			\textbf{0.1} & \textbf{0.6} & 4.37 \\
			
			& w/o ON & 0.9 & 0.0 & 0.1 & 0.4 & 0.0 & \textbf{0.1} & 0.2 & 4.73 \\
			& w/o VN & 0.7 & 0.0 & 0.2 & 0.5 & 0.0 & 0.0 & 0.1 & 6.24 \\
			\bottomrule
		\end{tabular}
	}
	
	\begin{tablenotes}
		\item[a] ``w/o ON" = without object nodes, ``w/o VN" = without viewpoint nodes.
	\end{tablenotes}
\vspace{-5mm}
\end{table}
\begin{figure}[b]
	\vspace{-5mm}
	\centering
	\includegraphics[width=0.99\linewidth]{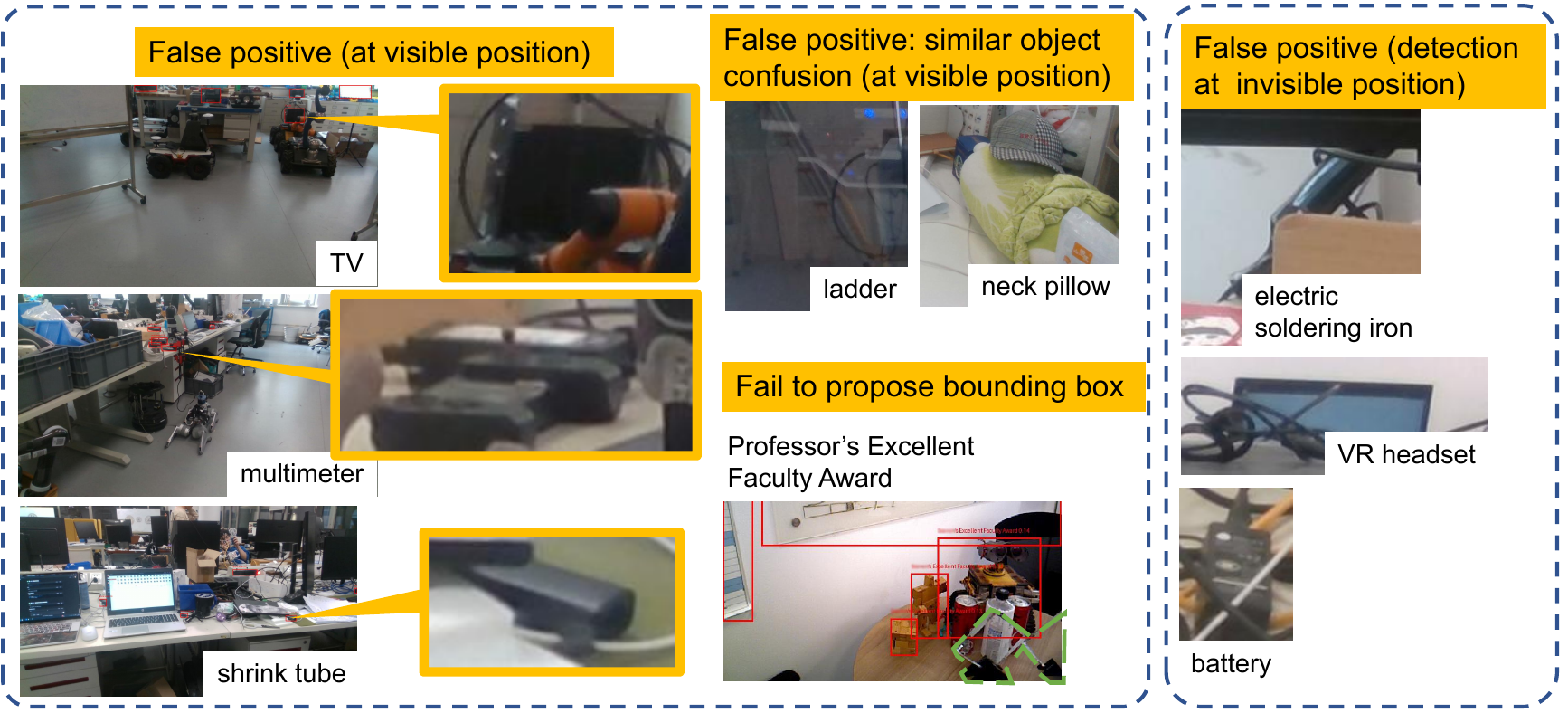}
	\vspace{-3mm}
	\caption{Our real-world experiments encountered nine failure cases, split by whether the robot could see the objects at those nodes.
		Six occurred at positions where the queried object was visible, including three detection errors, two confusion cases (misidentifying a ladder reflection as a ladder and a hugging pillow as a neck pillow), and one failure to generate bounding box for a transparent object.
		The other three failures resulted from false positive detections at non-visible nodes.}
	\label{fig:failed}
	\vspace{-6mm}
\end{figure}
\subsubsection{Failure Analysis}
Our experiments resulted in 9 failed cases out of 30 trials, as shown in Fig. \ref{fig:failed}.
In six cases (displayed in the first two columns of Fig. \ref{fig:failed}), the robot reached positions where the target objects were visible but still failed to retrieve them.
The remaining three failures happened when the robot stopped at nodes where the queried objects weren't visible, yet the detection method incorrectly reported positive identifications. 
Therefore, the system achieved 90\% correct robot positioning (27/30), with detection errors accounting for most failures.



\vspace{-0mm}

\section{Conclusion and Future Work}
In this letter, we propose the osmAG-LLM framework that combines a lightweight, text-based semantic map with LLM reasoning and active object search to perform object-goal navigation.
We show that this approach outperforms established, more high-fidelity mapping approaches and specifically shines when retrieving relocated or unmapped objects.
While these two cases are already more realistic, steps for our future research include searching for objects in concealed spaces such as drawers, as well as adding the ability to grasp the objects for full retrieval.

\bibliography{Bibliography}
\end{document}